# Design and development a children's speech database

*Radoslava Kraleva*

*South-West University "Neofit Rilski", Blagoevgrad, Bulgaria*

**Abstract**: The report presents the process of planning, designing and the development of a database of spoken children's speech whose native language is Bulgarian. The proposed model is designed for children between the age of 4 and 6 without speech disorders, and reflects their specific capabilities. At this age most children cannot read, there is no sustained concentration, they are emotional, etc. The aim is to unite all the media information accompanying the recording and processing of spoken speech, thereby to facilitate the work of researchers in the field of speech recognition. This database will be used for the development of systems for children's speech recognition, children's speech synthesis systems, games which allow voice control, etc. As a result of the proposed model a prototype system for speech recognition is presented.

**Keywords:** Children's speech recognition, Speech Corpora, Children's speech recognition system.

## 1. INTRODUCTION

The design and realization of the speech database is an integral part of systems for speech recognition. Their quality and scope greatly affect the activities of recognized devices [12]. Therefore, each acoustic-phonetic database must contain in itself all the phonetic richness of language studies. In our case the target group is children at the age of 4 to 6 (kindergarten and preschool group) whose native language is Bulgarian. Often they still cannot read and write, and a few of them know the numbers.

In practice, the collection of spoken speech of such speakers has proved a difficult task and additional resources had to be used. For example, the word to be pronounced is read, and then the child repeats it [4]. Often the selected words appear to be too complicated and the children have difficulty with their reproduction. In addition, young children easily lose concentration and distract, which further hampers the collection of data. Therefore, the purpose of this article is to collect a database that allows recording of additional helpful information related to supporting the process of filling a speech database of spoken speech.

For the development of the model methods that include selecting appropriate texts for spelling will be used, as well as visualization of these texts with images, moving images or sound files, labelling the recordings of each speaker and the organization of data in an easy and accessible form. Since such a research on children's recognition of speech in Bulgarian has not been done yet, the initial filling of the base will consist of only single words.

Such a database of recorded children's speech is a key condition for the application, starting from the pure commercialization as entertainment, education and other social and economically important areas. The resulting database will help many professionals, researchers and speech therapists involved with the study of children's speech.

## 2. ANALYSIS OF THE PROBLEM

According to [2] spoken speech databases are developed for two main goals - the first is to conduct fundamental research on the acoustic, phonetic, lexical, semantic, syntactic expressions of a language, and the second is to establish the differences between speakers, such as gender, age, environment, channels of data, etc.

In the development of systems related to the processing and speech recognition what is essential is the organization, flexibility and the size of the database. Also, the extent of actual coverage of the structure (syntax, word order, grammar, etc.) used in language and its phonetic features is important [9]. The first factor related to the design of the database determines the response time during the speech recognition and the linguistic factor controls the specificity of the speech.

It is also well known that children's speech contains a set of specific parameters [14], which turn them into a group of users with specific system requirements for automatic speech recognition. Frequent speech disorders are another specific problem to be solved [5]. At present the main trend in the data collection of spoken speech (speech data) of children at the age of 4 to 6 is the collecting and analyzing the acoustic and linguistic characteristics.

Most databases containing a record of children's spoken speech are targeted at children between the age of 6 and 18 (or their subset). This is because with this group the collection of speech is more easily manageable and thus feasible. They are primarily focused on acoustic modeling and analysis of American English as the corpora CID [14], KIDS [6], CU Kids' Audio Speech Corpus [8]. The most famous European corpus is PF-STAR, which contains spoken speech in British English, Italian, German and Swedish [3]. Example of the Russian spoken corpus

42

of children's speech is ChildRu [15]. Research and comparative characteristics of the existing corpora is done in [11]

Currently there are very few databases of spoken Bulgarian. One is developed with the project "Computer recognition of connected speech in a large dictionary of Bulgarian language" [16]. The read speech consists of general economic, legal and administrative texts, and the speakers are men and women aged over 20. The other database is BG-SRDat presented in [17]. It contains male speech (men at the age over 27) delivered Through a noisy analog telephone channels.

At the time of the survey and writing this article no children's spoken database in Bulgarian was found.

## 3. PREPARATION FO DEVELOPING THE DATABASE OF SPOKEN SPEECH

There are three main aspects that must be followed when designing a database of spoken speech - the type of dictionary, the number of sessions conducted with a speaker and technical aspects of recording [Lamel et al., 1986].

### *3.1. Dictionary*

According to [7] to achieve satisfactory results in the recognition process with a recognition system, the training must be done by the same target group, whose speech will be recognized. Therefore, records must be oriented towards young children (aged 4 to 6) which provides quick and convenient recording, with an appropriate balance between sex and age. Typically, the dictionary covers the full range of phonemes of the language. Here we face difficulties in young children because they use vocabulary different from that of adults. Therefore we will use suggestions from [1] frequency dictionary. The speakers used have to cover all sections of the existing speech-specific features at this age.

### *3.2. The number of sessions*

Under session we will understand all entries made by a speaker in a given period of time (e.g. per 1 day). This defines the length of the records and the number of sessions for each child who participated in the study. Records must be as short as possible. Each record contains a speech by one speaker.

### 3.3. . Technical aspects

They contain a representation of the environment, technical equipment (sound card, microphone type etc.) and algorithms for cleaning up the noise. Speech must be collected in a realistic way.

In our case records will be made at home, because from a psychological perspective, it is found that children at this age feel most relaxed and can more easily communicate with the conductor of the experiment. Each record will contain only one spoken word

## 4. MODEL OF CHILDREN'S SPOKEN SPEECH DATABASE

Most speech recognition systems don't allow the use of interactive elements during the recording of spoken language, such as moving objects, changing the image, listening to different sounds. This system is an attempt to remedy this shortcoming.

A relational diagram of the proposed database is presented in Fig. 1. The three main points on which emphasis falls are: the corpus (table" Words"), speaker (table "Speaker") and conducted recordings (table "Records").

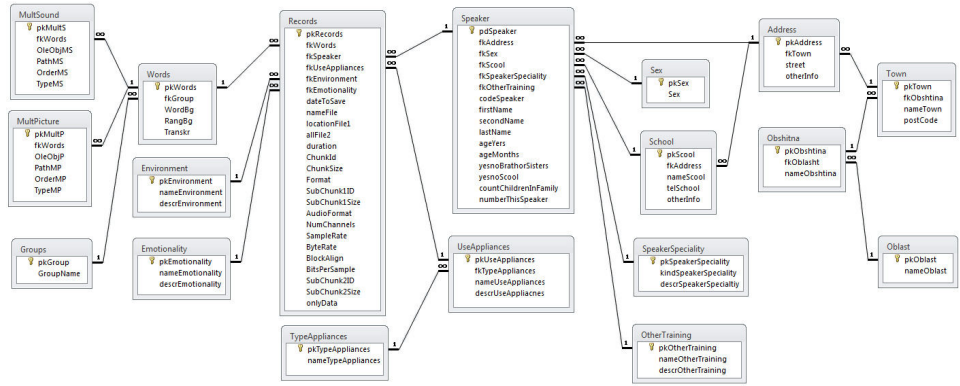

Fig. 1: Relational database diagram for children's speech

Let us now turn our attention to the information stored for each speaker, children whose speech will be collected. It consists of: full name, gender, age (years and months), date of birth, current address, number of children in family, number of the child in order of birth, attending a kindergarten, and if yes what kindergarten, whether attending any additional courses (speech therapist, singing lessons, music lessons), deviations from normal development and diseases.

The information collected about the records is the place of recording, recording equipment, emotional state of the speaker (the child) during the recording and the characteristics of the file. The recorded files are of low compression and extension wav. The original files can contain external noise, the voice of the mother or the tutor of the experiment, the child's laughing or crying, etc. Therefore, after its recording, filtering to eliminate the additional interference will be made.

## 5. EXPERIMENT

The experimental part consists of developing a prototype model of the proposed speech database (Fig. 2). There is a main form of which through the menu bar, the other modules are available. The software makes it easy to add new words to the database, there is an automatic phonetic transcription consistent with the International Phonetic Alphabet and application of illustrations and sound files to them. There is also an opportunity to work with electronic papers of the speakers.

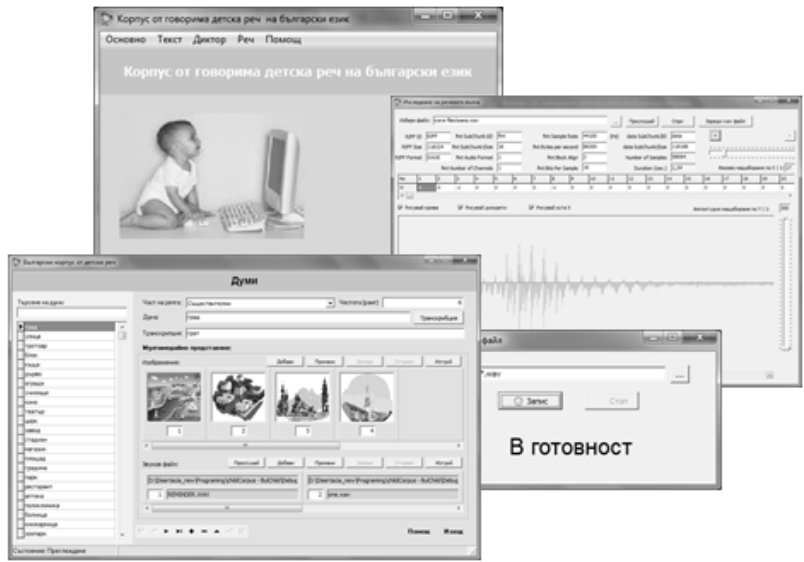

Fig. 2: View of the prototype system for identification children's speech in Bulgarian

Before proceeding to make records, first, we must organize the "Word's collection" for different sessions. They are organized thematically ("Seasons", "Family," "Numbers", etc.) and a small set of words (about 10-15 of them), since the young child cannot spend a long time (more

45

than 30 minutes a day) in front of the computer. Once the collections are ready one can continue with recording. From the main menu the form "Record" is started and the speaker is selected, the words collection, and in a dialog mode the words and the accompanying media files (pictures and sounds) are displayed. If the child copes with the first image and pronounces the necessary word correctly, the next picture follows and so on to the end.

All received files are uncompressed file format *. wav. The initial data are available in a simple file sampling rate 16 kHz, in a file channel. Sampling frequency (sample rate) shows the number of sample for one minute. Currently, signal filtering is done at the hardware level.

The file name is formed as a series of the letter b (boy) for a boy or g (girl) for a girl, a unique string (one or more letters), age, underscore, the collection, from which the word is, is marked with a letter and a serial number of the word (Tab.1).

Tab. 1. Example of a possible file name

| Attribute | Value | Indication |
| --- | --- | --- |
| Sex: | Boy | b |
| Name: | Ivan Petrov | A |
| Age: | 5 years | 5 |
| Word's collection: | Collection B | B |
| Word order: | Number of the word order 2 | 2 |
| Indication of the file: | | bA5_B2.wav |

Because children are highly dependent on the emotions that are directly reflected in speech and the manner of expression, then in the course of each record an opportunity to determine the emotional state of the child is provided. The first results for recognizing emotions in preschool children are presented in [10]. That is why records are made when children are visibly calm.

The presented prototype system was developed with the free development environment, Turbo C++ Explorer. Microsoft Access is used for the development of the database.

## 6. CONCLUSIONS

This report has completed several tasks. First, the process of developing a database of children's spoken language has been introduced and studied. Second, specific issues related to the recognition of child speech have been analyzed. And third, but no less important, architectural model of the speech database has been proposed. In the experimental part a prototype system for the children's recognition in Bulgarian has been presented.

The important advantage of this model is the narrow focus and comprehensive multimedia presentation of individual words in the corpus. In this way children's speech can be recorded with minimal involvement of the tutor.

This database is the beginning of developing a prototype system for automatic speech recognition of children at the age of 4 to 6 whose native language is Bulgarian. In the future the problem of rapid access to the contents of large multimedia databases and to explore the acoustic-phonetic diversity of existing database should be solved. The development of models for segmentation of phonemes from the words of recorded speech and calculate the word error rate is to be elaborated.